\documentclass[sigconf]{acmart}
\usepackage[table]{xcolor}
\usepackage{framed}
\usepackage{multirow}
\usepackage{booktabs}
\usepackage{graphicx}
\usepackage{pifont}

\AtBeginDocument{%
  }

\setcopyright{cc}
\setcctype{by}            
\copyrightyear{2026}
\acmYear{2026}
\setcopyright{cc}
\setcctype{by}
\acmConference[MM '26]{Proceedings of the 34th ACM International Conference on Multimedia}{November 10--14, 2026}{Rio de Janeiro, Brazil}
\acmBooktitle{Proceedings of the 34th ACM International Conference on Multimedia (MM '26), November 10--14, 2026, Rio de Janeiro, Brazil}
\acmDOI{10.1145/3767308.3835980}
\acmISBN{979-8-4007-2213-4/2026/11}

\makeatletter
\newdimen\acm@gap
\newcount\acm@rowcnt
\let\ACM@orig@author@bx\@typeset@author@bx
\def\@typeset@author@bx{%
  \global\advance\acm@rowcnt by 1\relax
  \ifnum\acm@rowcnt<4\relax
    \author@bx@wd=\dimexpr(\textwidth-2\acm@gap)/3-2pt\relax
  \else\ifnum\acm@rowcnt<8\relax
    \author@bx@wd=\dimexpr(\textwidth-3\acm@gap)/4-2pt\relax
  \else
    \author@bx@wd=\dimexpr(\textwidth-2\acm@gap)/3-2pt\relax
  \fi\fi
  \ACM@orig@author@bx
  \ifnum\acm@rowcnt=3\relax
    \unskip\break
  \else\ifnum\acm@rowcnt=7\relax
    \unskip\break
  \fi\fi
}
\author@bx@sep=\acm@gap
\makeatother

\begin{document}

\title{SpaAct: Spatially-Activated Transition Learning with Curriculum Adaptation for Vision-Language Navigation}

\author{Pengna Li}
\authornote{Both authors contributed equally to this research.}
\email{sauerfisch@stu.xjtu.edu.cn}
\affiliation{%
  \institution{Xi'an Jiaotong University}
  \city{Xi'an}
  \state{Shaanxi}
  \country{China}
}

\author{Kangyi Wu}
\authornotemark[1]
\email{wukangyi747600@stu.xjtu.edu.cn}
\affiliation{%
  \institution{Xi'an Jiaotong University}
  \city{Xi'an}
  \state{Shaanxi}
  \country{China}
}

\author{Shaoqing Xu}
\authornote{Project leader.}
\email{shaoqing.xu@connect.um.edu.mo}
\affiliation{%
  \institution{University of Macau}
  \city{Macau}
  \country{China}
}
\affiliation{%
  \institution{Xiaomi EV}
  \city{Beijing}
  \country{China}
}

\author{Fang Li}
\email{lifang7@xiaomi.com}
\affiliation{%
  \institution{University of Macau}
  \city{Macau}
  \country{China}
}
\affiliation{%
  \institution{Xiaomi EV}
  \city{Beijing}
  \country{China}
}

\author{Hanbing Li}
\email{lihanbing@xiaomi.com}
\affiliation{%
  \institution{Xiaomi EV}
  \city{Beijing}
  \country{China}
}

\author{Lin Zhao}
\email{zhaolins@foxmail.com}
\affiliation{%
  \institution{Beijing Institute of Technology}
  \city{Beijing}
  \country{China}
}

\author{Kailin Lyu}
\email{lvkailin2024@ia.ac.cn}
\affiliation{%
  \institution{Chinese Academy of Sciences}
  \city{Beijing}
  \country{China}
}

\author{Long Chen}
\email{alwaysunny@gmail.com}
\affiliation{%
  \institution{Xiaomi EV}
  \city{Beijing}
  \country{China}
}

\author{Zhi-Xin Yang}
\email{zxyang@um.edu.mo}
\affiliation{%
  \institution{University of Macau}
  \city{Macau}
  \country{China}
}

\author{Nanning Zheng}
\correspondingauthor
\email{nnzheng@mail.xjtu.edu.cn}
\affiliation{%
  \institution{Xi'an Jiaotong University}
  \city{Xi'an}
  \state{Shaanxi}
  \country{China}
}

\renewcommand{\shortauthors}{Pengna Li et al.}

\begin{abstract}
  Vision-and-Language Navigation (VLN) aims to enable an embodied agent to follow natural-language instructions and navigate to a target location in unseen 3D environments. Recent VLN methods increasingly build on general vision-language models (VLMs) and directly learn an end-to-end mapping from instructions and egocentric observations to low-level actions. While effective, this paradigm still suffers from a key limitation: 
  VLMs tend to learn priors that favor semantic understanding, making direct observation-to-action learning prone to learn shallow pattern matching rather than the dynamic spatial awareness that understands \textbf{how} and \textbf{why} the observation transition occurs over time. 
  We argue that adapting VLMs to VLN requires endowing them with two complementary capabilities for acquiring such awareness, namely backward action reasoning~(why) and forward transition prediction~(how).
  Based on this insight, we propose \textbf{SpaAct}, a simple yet effective training framework that activates the dynamic spatial awareness in VLMs. Specifically, SpaAct introduces two spatial activation tasks: \textbf{Action Retrospection}, which asks the model to infer the executed action sequence from visual transitions, and \textbf{Future Frame Selection}, which forces the model to predict the visual transitions conditioned on history and action.
  These two objectives provide lightweight supervision on both backward action reasoning and forward transition prediction, encouraging the model to build 
  dynamic spatial awareness
  in a VLM-friendly way. To further stabilize adaptation, we design \textbf{TriPA}, a \textbf{Tr}i-factor \textbf{P}rogressive \textbf{A}daptive curriculum learning method that organizes training samples from easy to hard,
  allowing the model to gradually acquire navigation skills from basic locomotion to long-horizon reasoning. Experiments on standard VLN-CE benchmarks show that SpaAct consistently improves VLM-based navigation and achieves state-of-the-art performance. We will release the code and models to support future research.
\end{abstract}

\ccsdesc[500]{Computing methodologies~Computer vision}
\ccsdesc[300]{Computing methodologies~Learning paradigms}
\ccsdesc[100]{Computing methodologies~Scene understanding}
\keywords{Vision-and-Language Navigation, Vision-Language Models, Spatial Understanding, Curriculum Learning}



\maketitle



\begin{figure*}
    \centering
    \includegraphics[
        width=0.98\textwidth,
    ]{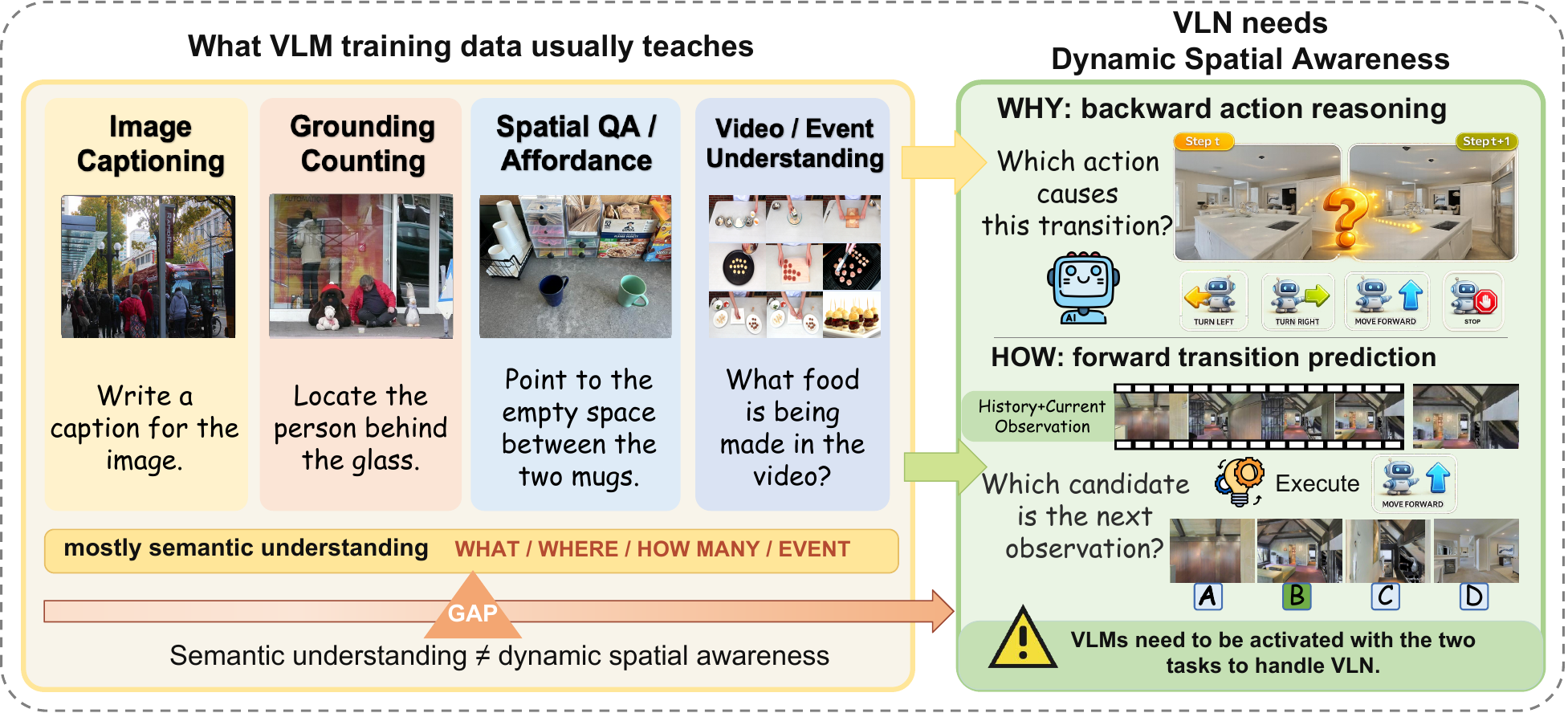}
    \caption{\textbf{Motivation of SpaAct.} Conventional VLM training mainly teaches semantic understanding, whereas VLN requires dynamic spatial awareness. SpaAct bridges this gap by activating backward action reasoning and forward transition prediction, yielding clear gains on VLN-CE benchmarks.}
    \label{fig:introduction}
\end{figure*}

\section{Introduction}

Vision-and-Language Navigation (VLN)~\cite{anderson2018vision,gu2022vision,guhur2021airbert,zhang2024vision} is a fundamental task in embodied artificial intelligence~\cite{li2026causalworldmodelingrobot,zhang2025embodied,chu2026abotn0technicalreportvla}, where an agent must follow natural-language instructions to navigate toward a target location in previously unseen 3D environments. Compared with classical navigation problems~\cite{sun2024fgprompt,yokoyama2024hm3d,li2025regnav} that rely only on geometric goals, VLN requires the agent to jointly understand language, perceive egocentric visual observations, and make sequential control decisions in a visually grounded manner. This makes VLN a representative benchmark for studying how general multimodal models can be adapted to real-world embodied decision-making.

Recent progress in VLN has been increasingly driven by vision-language models (VLMs)~\cite{bai2023qwen,bai2025qwen25vltechnicalreport,bai2025qwen3,lin2024vila}. Instead of relying on handcrafted modules~\cite{chen2025constraint,wang2023dreamwalker} or simulator-specific waypoint predictors~\cite{krantz2021waypoint,an2024etpnav}, many recent methods build on large pre-trained multimodal backbones and directly learn an end-to-end mapping from instructions and streaming observations to low-level actions~\cite{wei2025streamvln,cheng2024navila,zeng2025janusvln,zhang2025mapnav,huang2025mobilevla,liu2025navforesee}. This paradigm is attractive because it inherits strong semantic understanding, object recognition, and instruction following capabilities from VLMs, while also offering a simple and flexible interface for embodied control. 


Despite their strong semantic priors, current VLM-based VLN methods still face an important limitation. As illustrated in Fig.~\ref{fig:introduction}, although modern VLMs may also be exposed to videos or multi-frame data during pre-training, their training tasks mainly strengthen semantic understanding that focuses on what is or happens in the image or video.
However, in embodied navigation, the model needs to interact with 3D environments~\cite{chang2017matterport3d, ramakrishnan2021habitat}. Therefore, it's much more important to build dynamic spatial awareness that understands why the observation evolves over time and how the observation will change with the action executed. 

To help them build such awareness, we argue that it's required to equip VLMs with two complementary capabilities: \emph{backward action reasoning} and \emph{forward transition prediction}. Concretely, as presented in Fig.~\ref{fig:introduction}, a navigation-capable VLM should be able to reason backward from visual transitions to the actions that caused them, and also reason forward from the current state and action to the likely next observation. In this way, 
it explicitly models how actions cause visual transitions and why visual transitions happen from actions, mitigating shallow pattern matching.


Based on this insight, we propose \textbf{SpaAct}, a simple yet effective transition-aware training framework for vision-language navigation, equipped with a curriculum adaptation strategy for stable VLM adaptation.
SpaAct introduces two spatial activation tasks to explicitly encourage dynamic spatial awareness. The first task, \textbf{Action Retrospection}, asks the model to infer the executed action sequence from visual transitions between observations. Instead of requiring explicit geometric regression such as pose prediction or transformation estimation, this task casts inverse dynamics reasoning into a language-centered prediction format that is naturally compatible with VLM pre-training. The second task, \textbf{Future Frame Selection}, asks the model to identify the correct next observation conditioned on navigation history and the current action. Compared with heavy pixel-level future generation, this formulation offers a lightweight discriminative alternative that still encourages the model to internalize forward environmental dynamics. These two tasks respectively supervise backward action reasoning and forward transition prediction, allowing the model to develop dynamic spatial awareness.

To further stabilize this adaptation process, we introduce \textbf{TriPA}, a \textbf{Tr}i-factor \textbf{P}rogressive \textbf{A}daptive curriculum learning strategy. Training a general-purpose VLM for embodied navigation is inherently difficult because navigation samples vary substantially in trajectory length, instruction complexity, and motion patterns. Directly exposing the model to all training instances with equal probability leads to unstable optimization.
TriPA addresses this issue by organizing training samples from easy to hard according to trajectory, instruction, and motion complexity. In this way, the model can first acquire simple locomotion and short-horizon transition understanding, and then gradually develop more robust long-horizon reasoning abilities. 
TriPA complements the spatial activation tasks and makes training more stable and effective.

Our main contributions are summarized as follows:
\begin{itemize}
    \item We identify the spatial adaptation gap of VLM-based VLN and argue that robust embodied navigation requires activating VLMs with both backward action reasoning and forward transition prediction. 
    \item We propose \textbf{SpaAct}, a spatial-aware training framework that introduces two spatial activation tasks, \textbf{Action Retrospection} and \textbf{Future Frame Selection}, to encourage VLMs to learn dynamic spatial awareness in a lightweight and VLM-compatible manner.
    \item We design \textbf{TriPA}, a tri-factor progressive adaptive curriculum learning strategy that gradually organizes navigation samples from easy to hard,
    stabilizing optimization and promoting progressive acquisition of navigation skills.
    \item We conduct extensive experiments on VLN-CE benchmarks and show that SpaAct consistently improves navigation capability and achieves state-of-the-art performance.
\end{itemize}

\section{Related Works}
\subsection{Vision-Language Navigation}
Vision-language navigation (VLN) studies how an embodied agent follows natural language instructions to reach a target location in unseen environments~\cite{anderson2018vision,lyu2026instruction}. 
Earlier VLN methods mainly focused on discrete settings, where navigation is simplified into selecting high-level actions over predefined candidates~\cite{qiao2022hop,liu2023bird,gao2023adaptive,liu2026instruction,liu2026essence}. 
Recently, more works~\cite{krantz2020beyond,raychaudhuri2021language,hong2022bridging,irshad2022semantically,wang2023dreamwalker, an2024etpnav,yao2025navmorph} have begun to focus on vision language navigation in continuous environments~(VLN-CE).
To reduce control difficulty, many methods adopt hierarchical designs with simulator-trained waypoint predictors or other intermediate action abstractions~\cite{krantz2022sim,wang2023dreamwalker,an2024etpnav,li2025ground,lyu2026himemvln}. Nevertheless, due to their reliance on panoramic sensors, they often struggle to generalize to unseen scenes. Recently, vision language models~(VLMs)~\cite{bai2025qwen3,achiam2023gpt, zhang2024llava,} have achieved significant advances in multimodal understanding and reasoning, which has catalyzed a paradigm shift in vision language navigation. 
Many works~\cite{qi2025vln,cheng2024navila,zeng2025janusvln,wang2024sim, liu2025nav, zhang2025embodied, li2026think, wei2025streamvln} utilize VLMs as a backbone to perceive and understand visual observations and instructions, and then reason the next low-level actions, empowering agents with sophisticated reasoning capabilities. 
Despite the impressive progress made by the aforementioned methods, most of them focus on architectural modifications.
However, a significant domain gap remains between the semantic understanding acquired during VLM pre-training and the dynamic spatial awareness required for robust navigation. We argue that the bottleneck of current VLM-based agents is the misalignment between "knowing what" (semantics) and "knowing where/how" (spatial dynamics). 
Another line of work strengthens navigation with more explicit spatial structures, such as top-down semantic maps~\cite{georgakis2022cross}, occupancy-based spatial priors~\cite{liu2026span}, , explicit spatial perception and exploration~\cite{zeng2025janusvln,yue2026spatial}, or spatial scene graphs for zero-shot navigation~\cite{zhang2026spatialnav}. 
These methods improve spatial grounding through dedicated spatial representations or reasoning modules. In contrast, our method is map-free and lightweight: rather than adding explicit spatial structures at inference time, we adapt a general-purpose VLM through transition-aware training objectives to improve dynamic spatial awareness.

\subsection{Auxiliary Task Learning for VLN}
Auxiliary task learning has been widely explored in VLN to improve data efficiency and navigation robustness. Some works focused on temporal navigation progress~\cite{ma2019self, zhu2020vision,ma2019regretful,chen2025constraint, wang2025progress} and instruction alignment~\cite{fried2018speaker,hong2020sub,wu2026dual} to ensure that the agent maintains a consistent understanding of its navigation state. For VLM-based VLN, some works~\cite{cheng2024navila, wei2025streamvln} adopt extra multi-modal datasets~\cite{zhang2024llava, azuma2022scanqa,zhu2023multimodal}, massive synthetic environments~\cite{wang2023scaling,wei2025ground} or real-world human navigation videos~\cite{lin2023learning} to adapt to diverse visual distributions. Recently, the research focus of VLN has shifted toward advanced foresight and reasoning alignment. Some works~\cite{liu2025navforesee, yao2025navmorph, perincherry2025visual} introduce future-view modeling to empower agents to imagine the potential future observations. Reasoning-based methods~\cite{liu2025nav} employ step-by-step Chains-of-Thought (CoT) before output actions to enhance the navigation robustness. 
Despite these advances, Most prior efforts enhance VLN through stronger reasoning frameworks, additional data, or future-view modeling. They still pay less attention to how a VLM can be progressively activated for navigation-relevant spatial understanding. In contrast, we focus on progressively activating VLMs' spatial understanding ability for navigation through auxiliary transition-aware objectives together with curriculum learning, enabling the model to move from general-purpose multimodal understanding towards action-conditioned spatial perception and embodied decision making.

\begin{figure*}[tb]
    \centering
    \includegraphics[width=\textwidth]{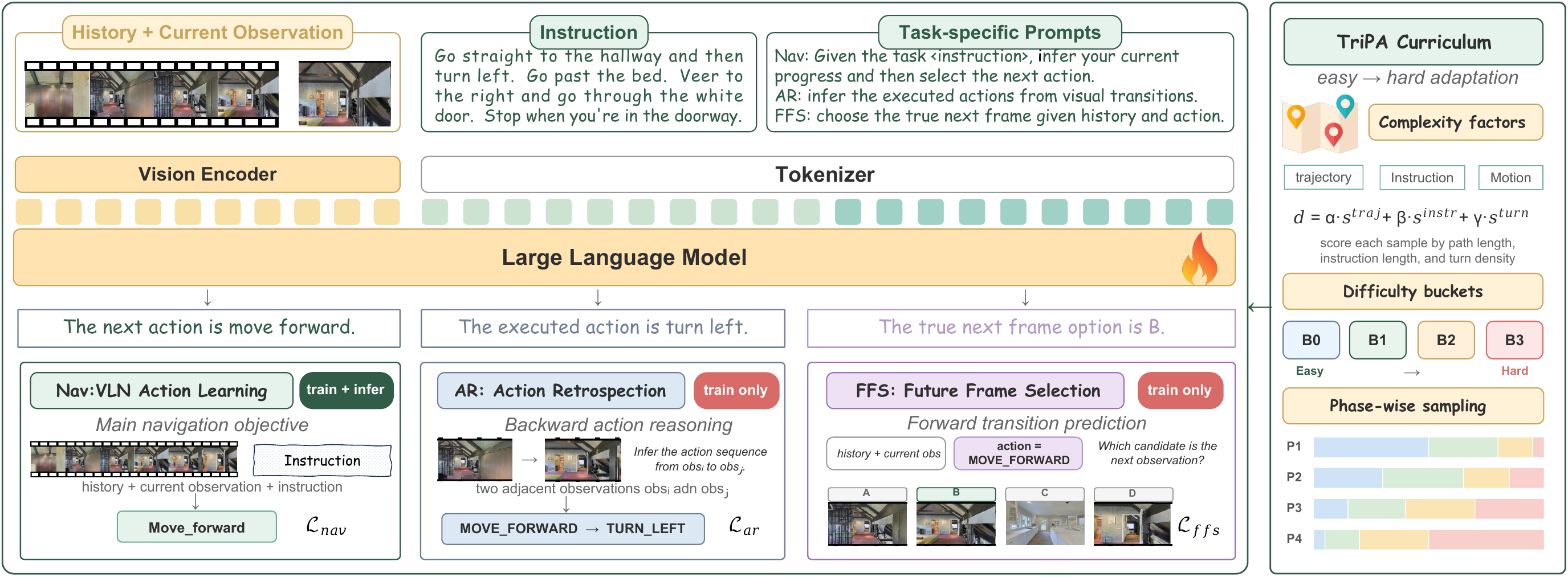}
    \caption{Overview of Our SpaAct. Given the instruction and the history and current observations, the VLM backbone is trained with the main navigation objective together with Action Retrospection and Future Frame Selection, while TriPA organizes training samples with a tri-factor easy-to-hard curriculum. During inference, only the navigation branch is used.}
    \label{fig:framework}
\end{figure*}

\subsection{Curriculum Learning for Embodied Navigation}
Curriculum learning has been explored in embodied AI as a training strategy that gradually exposes agents to easier, more informative, or better-structured experiences before harder ones. In VLN, Zhang et al.~\cite{zhang2021curriculum} showed that explicitly modeling sample difficulty and reorganizing training data with an easy-to-hard curriculum can improve navigation performance, generalization, and training efficiency without modifying the agent architecture. Beyond VLN, curriculum design has also been adopted in broader embodied navigation settings. For example, NavACL~\cite{morad2021embodied} introduces an automatic curriculum for visual navigation that selects training tasks based on geometric properties and outperforms uniform task sampling in reinforcement learning. 
Other works~\cite{zhang2023good,zhang2023robustness} use curricula over supervision availability rather than sample difficulty: train embodied navigation agents under progressively reduced teacher assistance. 
Different from these works, our curriculum is not a generic optimization strategy over navigation samples. Instead, TriPA is designed for VLM adaptation in VLN, where sample difficulty is defined from transition-related factors and used to progressively activate navigation-relevant ability.

\section{Method}
In this section, we present a unified framework for improving vision-and-language navigation through activating VLM with dynamic spatial awareness. We begin by formulating the problem and introducing the notations used in this work. Next, we develop a series of spatial activation tasks that explicitly encourage dynamic spatial awareness. Based on these tasks, we design a curriculum learning paradigm that organizes training samples in a progressive manner, enabling more robust optimization. Finally, we detail the training and inference procedures of the proposed approach. An overview of the approach is provided in Fig.~\ref{fig:framework}. 
\subsection{Preliminary}
\paragraph{Task Definition.} Vision-and-Language Navigation in continuous environments~(VLN-CE) can be formulated as a partially observable Markov decision process, where an embodied agent must navigate a 3D scene to reach a goal specified by a natural language instruction. Formally, given an text instruction $\mathcal{I} = \{w_1, w_2, \dots, w_L\}$ consisting of $L$ words, the agent is initialized at a starting location and must execute a sequence of actions to reach the target destination. At each time step $t$, the agent receives a monocular RGB observation $o_t \in \mathbb{R}^{H \times W \times 3}$, representing its current egocentric view. Based on the instruction $\mathcal{I}$, the current observation $o_t$, and the navigation history observations $\mathcal{H}_t$, the agent predicts a low-level action $a_t$ from a continuous-environment action space $\mathcal{A} = \{\text{\texttt{move\_forward, turn\_left, turn\_right, stop}}\}$. Each low-level action must corresponds to fine-grained control (e.g., moving $0.25m$, rotating $15^{\circ}$ or stop). The episode terminates when the agent issues the \texttt{stop} command or reaches a maximum step limit $T$. Success is defined by whether the final position $p_T$ is within a distance threshold $d_{success}$ from the ground-truth target $p^*$.

\paragraph{Backbone.}
We follow JanusVLN~\cite{zeng2025janusvln} and adopt a VLM-based navigation architecture as our backbone. Given the language instruction $\mathcal{I}$, current egocentric observation $o_t$, and navigation history $\mathcal{H}_t$, the model autoregressively predicts the next low-level action:
\begin{equation}
p_{\theta}(a_t \mid \mathcal{I}, \mathcal{H}_t, o_t)
= \mathrm{LLM}\big(\mathbf{E}(\mathcal{I}), \mathbf{E}(\mathcal{H}_t, o_t)\big),
\label{eq:janus_inference}
\end{equation}
where $\mathbf{E}$ denotes the embedded tokens of the instruction, history, and current observation. 

However, directly fine-tuning a pre-trained VLM for VLN still suffers from a spatial adaptation gap, as the model does not naturally capture ego-motion dynamics and action-conditioned scene transitions. To bridge this gap, we introduce two spatial activation tasks, \emph{Action Retrospection} and \emph{Future Frame Selection}, together with a curriculum learning strategy that progressively adapts the model from simple locomotion to more complex long-horizon navigation.

\subsection{Spatial Activation Tasks}

\subsubsection{Action Retrospection}
To empower general VLMs with the capability to navigate effectively in 3D environments, it is essential to foster an inherent understanding of 3D consistency~\cite{zhan2025actial}. 
In essence, 3D consistency means that objects in the environment remain geometrically coherent when observed from different viewpoints, even though their 2D projections may change with camera movement. This property is fundamental for embodied navigation, because it allows the model to connect visual observations across time and viewpoints, and thus supports spatial reasoning over the environment.
However, directly asking a general VLM to predict precise 6-DoF camera poses or relative transformation matrices from multi-view observations is highly challenging. Such outputs are continuous, geometry-heavy, and far removed from the language-oriented pre-training objectives of most VLMs. As a result, this formulation introduces a substantial gap between the desired spatial reasoning capability and the model's native prediction space.
To reduce this gap while still encouraging spatial-temporal awareness, we reformulate the problem as a language-centered activation task, which we call the \emph{action retrospection task}. Instead of requiring explicit geometric regression, this task guides the model to infer past actions from visual changes, allowing it to develop an implicit understanding of viewpoint transition and spatial consistency in a form that is more compatible with the original capabilities of VLMs.

\paragraph{Data Generation.}
The goal of Action Retrospection is to force the model to infer the underlying physical movement that caused the visual transition between two temporally ordered observations. Formally, given two observations sampled from a navigation trajectory, the model is tasked with predicting the executed actions that led from one to the other. We construct the retrospection dataset by extracting image pairs from navigation trajectories in R2R-CE~\cite{krantz2020beyond} and RxR-CE~\cite{ku2020room} training datasets, which can be formulated as:
\begin{equation}
\mathcal{D}_{ar} =\{o_i, o_j, A_e=\{a_i,...,a_{j-1}\} \}^{N_a},
\label{eq:action_retrospection}
\end{equation}
where $o_i$ and $o_j$ denote the sampled observations and $A_e$ is the executed actions. To ensure the model learns robust motion dynamics, we implement an action balancing strategy during data synthesis. This prevents the model from developing a majority-class bias toward moving forward and encourages it to focus on the subtle visual shifts associated with rotation. 

\paragraph{Training Procedure}
To train the VLM on the action retrospection task, we convert the sampled observation pairs and the ground-truth action sequence into a structured multi-modal prompt. The format is summarized as:
\begin{leftbar}
\noindent
{\small \textit{\textbf{User:}} \textit{\textbf{You will be given two observations from the same navigation trajectory. Infer the most likely action sequence that moves the agent from the first observation <image> to the second observation <image>. The predicted sequence must only use the following actions: MOVE\_FORWARD, TURN\_LEFT, TURN\_RIGHT, STOP.}}}
\end{leftbar}
\noindent The model takes a sequence of two images $(o_i, o_j)$ as input and is tasked with generating the corresponding action tokens. For each sample in $\mathcal{D}_{ar}$, we denote the instruction-based prompt as $P$ and the ground-truth action sequence as $A_e$. The training objective for this activation task is formulated as a next-token prediction loss:
\begin{equation}\mathcal{L}_{ar} = \mathbb{E}_{\mathcal{D}_{ar}}\big[-\sum \log p_{\theta}(\hat{A_e}=A_e \mid o_i, o_j, P)\big],\label{eq:loss_ar}\end{equation}
where $\hat{A_e}$ is the predicted actions. 
By training the VLM to "look back" and reason about its actions, we effectively inject an implicit inverse dynamics model into the backbone, enabling the agent to better perceive how its ego-motion transforms the 3D visual scene.

\subsubsection{Future frame selection}
While Action Retrospection enables the model to infer the causes of visual changes, a robust navigation agent also requires the ability to anticipate the consequences of its actions. This predictive capability is a hallmark of World Models, which allow agents to simulate potential future states internally. However, conventional pixel-level future frame generation is computationally expensive and potentially misaligned with the discrete, language-centric output space of VLMs. To maintain a lightweight activation mechanism while capturing environmental dynamics, we simplify the generative forward-modeling task into a discriminative Future Frame Selection task. 
\paragraph{Data Generation.}The objective of Future Frame Selection is to cultivate a "forward-thinking" capability within the VLM, enabling it to predict the visual consequences of its decisions. For each training sample, we provide the model with a history of observations $\mathcal{H}_t = \{o_{t-k}, \dots, o_{t-1}\}$, the current observation $o_t$, and the intended action $a_t$. The model must identify the ground-truth next observation $o_{t+1}$ from a set of four candidates $\mathcal{C} = \{c_A, c_B, c_C, c_D\}$. We construct the candidate set $\mathcal{C}$ using a hybrid negative sampling strategy to ensure task difficulty:
\begin{itemize}\item Same-trajectory Hard Negatives: we select observations from the same trajectory that are temporally close to the ground truth but do not correspond to the result of action. These act as hard distractors that force the model to distinguish subtle geometric shifts (e.g., rotating $15^{\circ}$ vs. $30^{\circ}$).
\item Cross-trajectory Negatives: we randomly sample observations from different trajectories to ensure the model maintains a broad discriminative boundary across diverse scenes.\end{itemize}
To ensure balanced learning across different motion types, we apply an action-balancing filter similar to the retrospection task. The dataset is formulated as:
\begin{equation}\mathcal{D}_{ffs} = \{ \mathcal{H}_t, o_t, a_t, \mathcal{C}, y \}^{N_f},\label{eq:future_frame_selection}\end{equation}
where $y \in \{A, B, C, D\}$ is the label of the correct option.
\paragraph{Training Procedure.} We formulate this task as a multiple-choice visual question-answering problem, which is highly compatible with the pre-trained capabilities of VLMs. The prompt includes the historical observations, followed by the current observation and the action. The format is summarized as:\begin{leftbar}\noindent{\small \textit{\textbf{User:}} \textit{\textbf{You are given a current observation <image>, your history observation <image>...<image>, and the executed action: [ACTION]. Which of the following candidates is the true next observation? Option A: <image>, Option B: <image>, Option C: <image>, Option D: <image>. Please select the correct option.}}}\end{leftbar}\noindent The model is trained to predict the correct option. For each sample in $\mathcal{D}_{ffs}$, we minimize the cross-entropy loss between the predicted logit and the ground-truth label $y$:
\begin{equation}\mathcal{L}_{ffs} = \mathbb{E}_{\mathcal{D}_{ffs}}\big[-\log p_{\theta}(\hat{y} = y \mid \mathcal{H}_t, o_t, a_t, \mathcal{C}, P)\big],\label{eq:loss_ffs}\end{equation}
where $P$ is the task-specific prompt. By optimizing this objective, the model effectively internalizes the forward dynamics of the environment, which encourages the agent to proactively anticipate environmental changes induced by its own actions. 

\begin{figure*}[t]
    \centering
    \includegraphics[width=\textwidth]{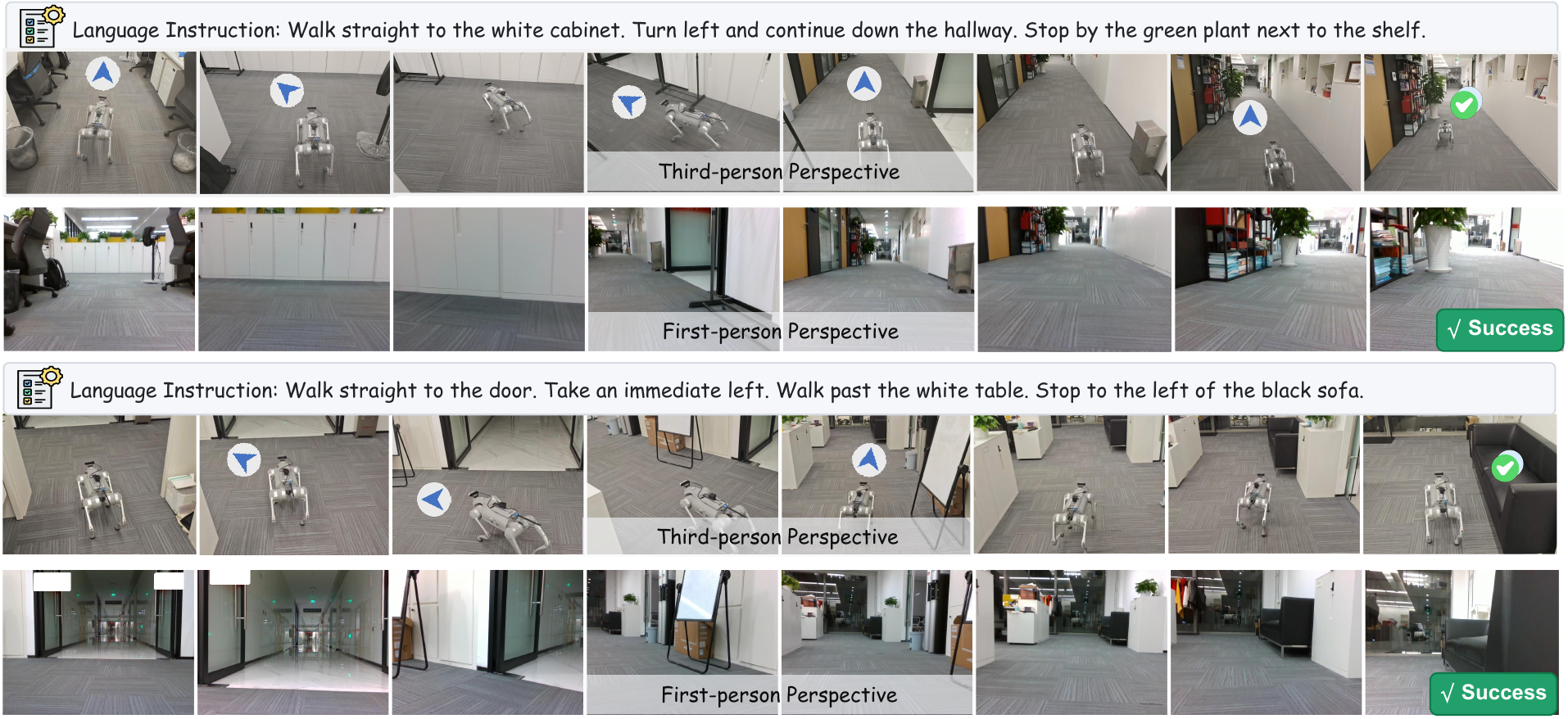}
    \caption{Real-world qualitative results of SpaAct on a Unitree Go2 robot. We provide both third-person views and egocentric observations. It's shown that the agent successfully follows the instruction and reaches the target location.}
    \label{fig:real_world}
\end{figure*}

\subsection{TriPA Curriculum Learning}
Training a general VLM for embodied navigation is inherently challenging due to the high variance in sample difficulty. If a model is exposed to complex, long-horizon trajectories prematurely, the resulting gradient instability can hinder the acquisition of basic navigation skills.
To address this, we introduce \textbf{TriPA} (\textbf{Tri}-factor \textbf{P}rogressive \textbf{A}daptive curriculum learning) to reshape the training distribution by presenting samples in a coarse-to-fine difficulty order, ensuring the VLM masters basic locomotion before tackling long-horizon spatial reasoning.

A core premise of our TriPA curriculum is that not all navigation tasks are created equal. Some require only simple forward movement, while others demand intricate multi-stage reasoning. To quantify this, we assign a scalar difficulty score $d_i$ to each sample $x_i$ based on three complementary factors:\begin{equation}d_i = \alpha s_i^{\mathrm{traj}} + \beta s_i^{\mathrm{instr}} + \gamma s_i^{\mathrm{turn}},\label{eq:difficulty_score}\end{equation}where $s_i^{\mathrm{traj}}$, $s_i^{\mathrm{instr}}$, and $s_i^{\mathrm{turn}}$ represent trajectory, instruction, and motion complexity, and $\alpha$, $\beta$, $\gamma$ denote the corresponding weights. 
For Trajectory Complexity ($s_i^{\mathrm{traj}}$), We use the normalized log-count of trajectory frames as a proxy for path length:
\begin{equation}
s_i^{\mathrm{traj}} = \text{Norm}\bigl(\log(1 + N_i^{\mathrm{png}})\bigr),
\end{equation}
where $N_i^{\mathrm{png}}$ denotes the number of images in the $i_{th}$ trajectory, reflecting the temporal window size.
For Instruction Complexity ($s_i^{\mathrm{instr}}$), we approximate the complexity of cross-modal grounding using the word count of the navigation instruction:
\begin{equation}
s_i^{\mathrm{instr}} = \text{Norm}(L_i^{\mathrm{instr}}),
\end{equation}
where $L_i^{\mathrm{instr}}$ represents the instruction length.
For Motion Complexity ($s_i^{\mathrm{turn}}$), we quantify the intensity of viewpoint transitions via turn density:
\begin{equation}
s_i^{\mathrm{turn}} = \frac{N_i^{\text{left}} + N_i^{\text{right}}}{N_i^{\text{all}}},
\end{equation}
where $N_i^{\text{left}}$ and $N_i^{\text{right}}$ are the counts of rotation actions, and $N_i^{\text{all}}$ is the total number of actions in the trajectory.
Then we sort all samples by $d_i$ and divide them into four difficulty buckets $\{\mathcal{B}_0, \dots, \mathcal{B}_3\}$. To avoid catastrophic forgetting of easy samples while introducing hard ones, we employ a phase-wise distribution shift. We define four training phases, each with a specific sampling weight vector~($\mathbf{P}{1} = [0.50, 0.30, 0.15, 0.05]$, $\mathbf{P}{2} = [0.35, 0.30, 0.25, 0.10]$, $\mathbf{P}{3} = [0.10, 0.25, 0.30, 0.35]$, $\mathbf{P}{4} = [0.05, 0.15, 0.30, 0.50]$) over the buckets. 
The final training order for an epoch is constructed by concatenating the shuffled samples of each phase. This stochastic bucketed approach provides a "soft" curriculum. It prevents the model from being overwhelmed by complex outliers in early stages while maintaining enough data diversity to avoid distribution collapse, which is critical for fine-tuning large-scale VLMs with high sensitivity to sample variance.

\subsection{Training and Inference}                 
We train SpaAct in two stages. In Stage~1, we jointly optimize the main navigation objective and the proposed spatial activation tasks. The training data consist of 0.69M Action Retrospection samples, 0.69M Future Frame Selection samples, and 2.4M navigation samples from R2R-CE~\cite{zhu2020vision} and RxR-CE~\cite{ku2020room}. The objective is
\begin{equation}
\mathcal{L}_{\mathrm{stage1}}=\mathcal{L}_{\mathrm{nav}}+\lambda_{ar}\mathcal{L}_{ar}+\lambda_{ffs}\mathcal{L}_{ffs}.
\end{equation}

In Stage~2, following JanusVLN~\cite{zeng2025janusvln}, we further improve the navigation policy with extra data. We collect 11.7K DAgger~\cite{ross2011reduction} trajectories from the Stage~1 model and further incorporate 155K trajectories from a subset of ScaleVLN~\cite{wang2023scaling}, which corresponding to \~10476k samples. These data are combined with the original 2.4M navigation samples from R2R-CE and RxR-CE, and we optimize only the navigation objective:
\begin{equation}
\mathcal{L}_{\mathrm{stage2}}=\mathcal{L}_{\mathrm{nav}}.
\end{equation}

During inference, the model directly predicts the next action conditioned on the instruction, current observation, and cached history. The spatial activation task and TriPA curriculum are used only during training, so inference remains lightweight. 

\begin{table*}[t]
\centering
\scriptsize
\setlength{\tabcolsep}{3pt}
\renewcommand{\arraystretch}{1.05}
\caption{Comparison with prior VLN-CE methods on the R2R-CE and RxR-CE val-unseen splits. SpaAct uses only a single RGB sensor, and external data denotes supervision beyond the standard R2R/RxR-CE training sets.}
\label{tab:r2r_rxr_sota}
\resizebox{\textwidth}{!}{
\begin{tabular}{lccccccccccccc}
\toprule
\multirow{2}{*}{Method} & \multicolumn{4}{c}{Observation} & \multicolumn{4}{c}{R2R Val-Unseen} & \multicolumn{4}{c}{RxR Val-Unseen} & \multirow{2}{*}{External Data} \\
\cmidrule(lr){2-5}\cmidrule(lr){6-9}\cmidrule(lr){10-13}
& Pano. & Odo. & Depth & S.RGB & NE$\downarrow$ & OS$\uparrow$ & SR$\uparrow$ & SPL$\uparrow$ & NE$\downarrow$ & SR$\uparrow$ & SPL$\uparrow$ & nDTW$\uparrow$ & \\
\midrule
Sim2Sim~(ECCV'22)~\cite{krantz2022sim}         & $\checkmark$ & $\checkmark$ & $\checkmark$ &              & 6.07 & 52.0 & 43.0 & 36.0 & -    & -    & -    & -    & - \\
VLN$\circlearrowright$BERT~(CVPR'22)~\cite{hong2022bridging}        & $\checkmark$ & $\checkmark$ & $\checkmark$ &              & 5.74 & 53.0 & 44.0 & 39.0 & 8.98 & 27.0 & 22.6 & 46.7 & - \\
Ego$^2$-Map~(ICCV'23)~\cite{hong2023learning}     & $\checkmark$ & $\checkmark$ & $\checkmark$ &              & 5.54 & 56.0 & 47.0 & 41.0 & -    & -    & -    & -    & - \\
DreamWalker~(ICCV'23)~\cite{wang2023dreamwalker}     & $\checkmark$ & $\checkmark$ & $\checkmark$ &              & 5.53 & 59.0 & 49.0 & 44.0 & -    & -    & -    & -    & - \\
GridMM~(ICCV'23)~\cite{wang2023gridmm}          & $\checkmark$ & $\checkmark$ & $\checkmark$ &              & 5.11 & 61.0 & 49.0 & 41.0 & -    & -    & -    & -    & - \\
Reborn~(ICCV'23)~\cite{an20221st}          & $\checkmark$ & $\checkmark$ & $\checkmark$ &              & 5.40 & 57.0 & 50.0 & 46.0 & 5.98 & 48.6 & 42.0 & 63.3 & - \\
InstructNav~(CoRL'24)~\cite{long2024instructnav}     & $\checkmark$ & $\checkmark$ & $\checkmark$ &              & 6.89 & -    & 31.0 & 24.0 & -    & -    & -    & -    & - \\
\midrule
COSMO~(ICCV'25)~\cite{zhang2025cosmo}           & $\checkmark$ &              &              &              & -    & 56.0 & 47.0 & 40.0 & -    & -    & -    & -    & - \\
AO-Planner~(AAAI'25)~\cite{chen2025affordances}      & $\checkmark$ &              & $\checkmark$ &              & 5.55 & 59.0 & 47.0 & 33.0 & 7.06 & 43.3 & 30.5 & 50.1 & - \\
MapNav~(ACL'25)~\cite{zhang2025mapnav}          &              & $\checkmark$ & $\checkmark$ & $\checkmark$ & 4.93 & 53.0 & 39.7 & 37.2 & -    & -    & -    & -    & - \\
g3D-LF~(CVPR'25)~\cite{wang2025g3d}          &              & $\checkmark$ & $\checkmark$ & $\checkmark$ & 5.70 & 59.5 & 47.2 & 34.6 & -    & -    & -    & -    & - \\
NaVid-4D~(ICRA'25)~\cite{liu2025vid}        &              &              & $\checkmark$ & $\checkmark$ & 5.99 & 55.7 & 43.8 & 37.1 & -    & -    & -    & -    & - \\
NavMorph~(ICCV'25)~\cite{yao2025navmorph}        &              &              & $\checkmark$ & $\checkmark$ & 5.75 & 56.9 & 47.9 & 33.2 & 8.85 & 30.8 & 22.8 & 44.2 & - \\
\midrule
NaVid~(RSS'24)~\cite{zhang2024navid}            &              &              &              & $\checkmark$ & 5.47 & 49.1 & 37.4 & 35.9 & -    & -    & -    & -    & 953K \\
Sim2Real~(CoRL'24)~\cite{wang2024sim}        &              &              &              & $\checkmark$ & 5.95 & 55.8 & 44.9 & 30.4 & 8.79 & 36.7 & 25.5 & 18.1 & 0K \\
StreamVLN*~(ICRA'26)~\cite{wei2025streamvln}     &              &              &              & $\checkmark$ & 6.05 & 53.8 & 45.5 & 41.6 & -    & -    & -    & -    & 10033K \\
Uni-NaVid~(RSS'25)~\cite{zhang2024uni}        &              &              &              & $\checkmark$ & 5.58 & 53.3 & 47.0 & 42.7 & 6.24 & 48.7 & 40.9 & -    & 3577K \\
NaVILA*~(RSS'25)~\cite{cheng2024navila}          &              &              &              & $\checkmark$ & 5.37 & 57.6 & 49.7 & 45.5 & -    & -    & -    & -    & 12574K \\
JanusVLN*~(ICLR'26)~\cite{zeng2025janusvln}        &              &              &              & $\checkmark$ & 5.17 & 58.0 & 52.8 & 49.2 & 6.46 & 51.4 & 44.3 & 59.1 & 0K \\
\rowcolor{gray!20}
\textbf{SpaAct-stage1 (Ours)}    &              &              &              & $\checkmark$ & \textbf{4.89} & \textbf{62.7} & \textbf{55.1} & \textbf{50.8} & \textbf{5.98} & \textbf{53.4} & \textbf{46.3} & \textbf{61.7} & \textbf{0K} \\
NaVILA~(RSS'25)~\cite{cheng2024navila}           &              &              &              & $\checkmark$ & 5.22 & 62.5 & 54.0 & 49.0 & 6.77 & 49.3 & 44.0 & 58.8 & 13132K \\
StreamVLN~(ICRA'26)~\cite{wei2025streamvln}      &              &              &              & $\checkmark$ & 4.98 & 64.2 & 56.9 & 51.9 & 6.22 & 52.9 & 46.0 & 61.9 & $\sim$26330K \\
JanusVLN(ICLR'26)~\cite{zeng2025janusvln}         &              &              &              & $\checkmark$ & 4.78 & 65.2 & 60.5 & 56.8 & 6.06 & 56.2 & 47.5 & 62.1 & 10692K \\
\rowcolor{gray!20}
\textbf{SpaAct-stage2 (Ours)}    &              &              &              & $\checkmark$ & \textbf{4.70} & \textbf{69.9} & \textbf{62.2} & \textbf{57.2} & \textbf{5.82} & \textbf{58.6} & \textbf{49.1} & \textbf{64.3} & \textbf{10476K} \\
\bottomrule
\end{tabular}
}
\end{table*}

\begin{table}[t]
\centering
\renewcommand{\arraystretch}{1}
\caption{Ablation study of TriPA, Action Retrospection (AR), and Future Frame Selection (FFS) on R2R-CE val-unseen. The full model achieves the best performance.}
\label{tab:ablation}
\setlength{\tabcolsep}{5pt}
\resizebox{0.95\columnwidth}{!}{
\begin{tabular}{ccc|cccc}
\toprule
\textbf{TriPA} & \textbf{AR} & \textbf{FFS} 
& \textbf{NE} $\downarrow$ 
& \textbf{OS} $\uparrow$ 
& \textbf{SR} $\uparrow$ 
& \textbf{SPL} $\uparrow$ \\
\midrule
~ & ~ & ~ 
& 6.27 & 53.05 & 46.02 & 42.24 \\

\checkmark & ~ & ~ 
& 5.80 & 54.76 & 48.86 & 45.77 \\

~ & \checkmark & ~ 
& 5.56 & 55.83 & 49.37 & 46.51 \\

~ & ~ & \checkmark 
& 5.16 & 54.15 & 47.06 & 44.07 \\

~ & \checkmark & \checkmark 
& 5.02 & 58.76 & 51.55 & 48.29 \\

\checkmark & \checkmark & \checkmark 
& \textbf{4.89} & \textbf{62.67} & \textbf{55.06} & \textbf{50.76} \\
\bottomrule
\end{tabular}
}
\end{table}

\section{Experiment}
\subsection{Experimental Setup}
\subsubsection{Dataset and Evaluation Metrics.}
We benchmark our method on the standard R2R-CE and RxR-CE datasets. Both datasets are instantiated within the Habitat simulator, utilizing high-fidelity indoor environment scans from Matterport3D. 
The validation unseen splits comprise 1,839 evaluation trajectories for R2R-CE~\cite{anderson2018vision} and 3,669 for RxR-CE~\cite{ku2020room}. The model's performance is quantified using a comprehensive suite of metrics: Success Rate (SR), Success-weighted Path Length (SPL), Oracle Success Rate (OS), Navigation Error (NE), and normalized Dynamic Time Warping (nDTW). Consistent with the broader continuous navigation literature~\cite{wei2025streamvln,zeng2025janusvln,li2025regnav}, we prioritize SR and SPL as the primary indicators, as they directly measure the agent's task completion and trajectory efficiency.

\begin{figure*}[t]
  \centering
  \includegraphics[width=0.94\linewidth]{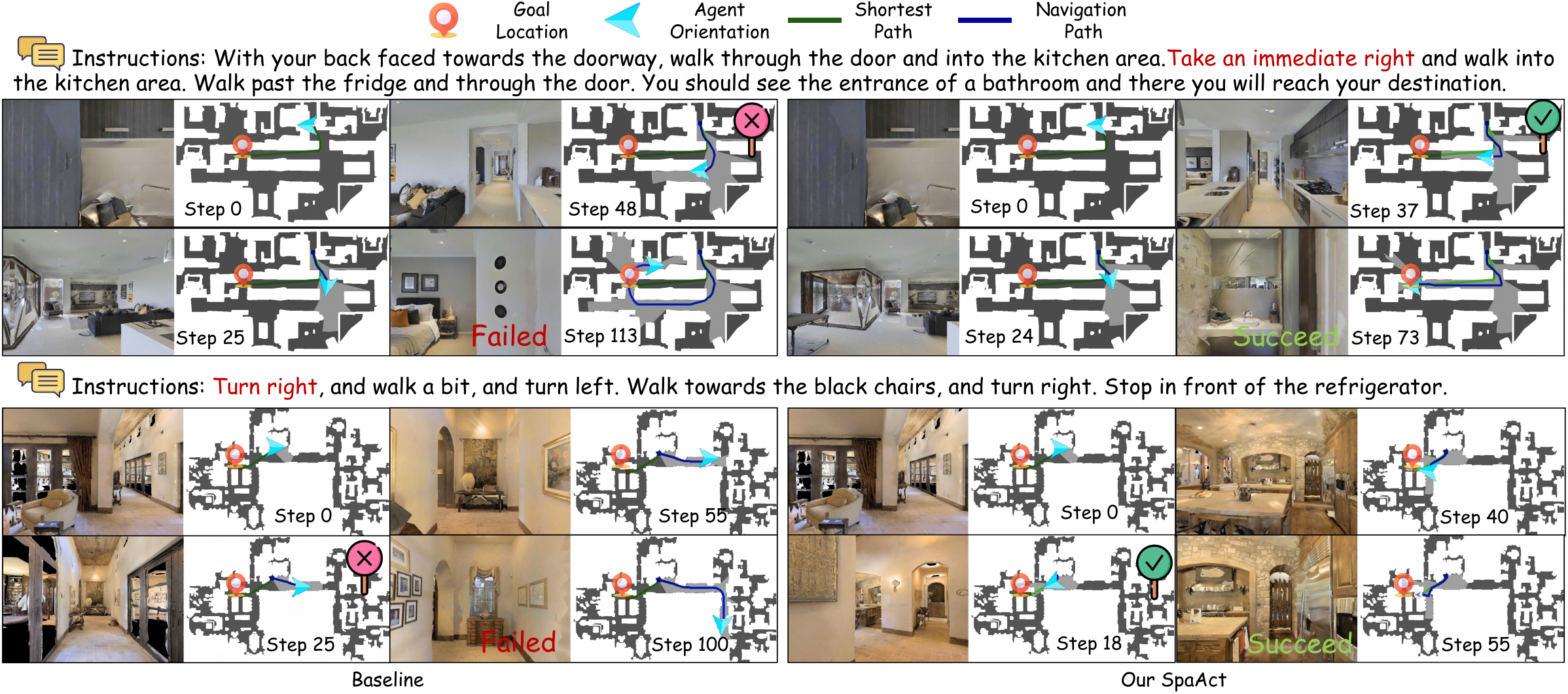}
  \caption{Qualitative comparison in two simulated unseen R2R-CE episodes. Compared with the baseline, SpaAct follows a more accurate trajectory and successfully reaches the goal, while the baseline deviates from the intended route and fails.}
  \label{fig:qualitative_sim}
  \vspace{-2mm}
\end{figure*}

\subsubsection{Implementation Details.}

Our implementation builds upon JanusVLN~\cite{zeng2025janusvln}, and we follow the default architecture and training setup of JanusVLN. For TriPA, we set the weights of trajectory complexity, instruction complexity, and motion complexity to $0.6$, $0.2$, and $0.2$, respectively. The loss weights for Action Retrospection and Future Frame Selection are set to $\lambda_{ar}=1$ and $\lambda_{ffs}=1$. For label balancing in the spatial activation tasks, we align all action categories to the size of the minority class. In Future Frame Selection, the history length is set to $4$ frames, while for navigation training data, the history length is set to $8$ frames.

\subsection{Main Results}
\subsubsection{Real-world Qualitative Results.}
We conduct real-world navigation experiments using a Unitree Go2 quadrupedal robot. The robotic platform is fitted with an Intel RealSense D435i RGB-D camera mounted on the front to capture egocentric visual observations. 
We adopt a client-server architecture to manage the computational workload during deployment. Specifically, continuous RGB video streams are transmitted from the robot to a remote server equipped with NVIDIA H20 GPUs. Once a natural language instruction is issued, the server-side model processes the incoming sensory data to infer the next discrete navigational action in real time. This action is subsequently sent back to the quadruped, which executes the physical movement via its native motion API. It is important to note that our navigation agent operates in a strict zero-shot manner; the model is trained entirely within the simulated Matterport3D environments~\cite{chang2017matterport3d} and directly deployed to physical spaces without any real-world fine-tuning or adaptation. Fig.~\ref{fig:real_world} provides two examples of real-world deployment of SpaAct. Although trained only in simulation, the model can still follow a multi-step real-world instruction and stop at the correct destination.


\subsubsection{Comparison with State-of-the-Arts on Simulated Environments.}
Table~\ref{tab:r2r_rxr_sota} compares SpaAct with prior methods on the R2R-CE and RxR-CE val-unseen splits. Using only a single RGB stream, SpaAct achieves strong generalization in unseen environments. 
In Stage 1, without any external data beyond the R2R/RxR-CE training sets, SpaAct-stage1 outperforms the 0K JanusVLN baseline, improving SR/SPL from 52.8/49.2 to 55.1/50.8 on R2R-CE and from 51.4/44.3 to 53.4/46.3 on RxR-CE. This shows that the proposed spatial activation objectives and TriPA alone can substantially improve VLM adaptation to embodied navigation. In Stage 2, after further training with external data~(Dagger data and Scalevln), SpaAct 
improves from 55.0/50.8 to 62.2/57.2 on R2R-CE and from 53.4/46.3 to 58.6/49.1 on RxR-CE, 
outperforming strong recent baselines under the same external-data setting. These results suggest that SpaAct benefits from three complementary components: Action Retrospection for backward action reasoning, Future Frame Selection for forward transition grounding, and TriPA for stable easy-to-hard training.

\subsubsection{Ablation Study of Components.}
Table~\ref{tab:ablation} verifies the contribution of each component in SpaAct. All experiments are done with SpaAct-stage1 to eliminate the effect of external data. AR and FFS improve over the baseline when applied individually.
Combining AR and FFS yields further gains, indicating that the two spatial activation tasks are complementary. Moreover, TriPA brings additional improvement on top of these objectives by stabilizing optimization through progressive easy-to-hard training. The full model achieves the best overall result, improving SR/SPL from 46.02/42.24 to 55.06/50.76, which suggests that SpaAct benefits from both spatial activation tasks and curriculum training strategy.

\subsubsection{Simulated Qualitative Results.}

Fig.~\ref{fig:qualitative_sim} provides qualitative comparisons between the baseline and SpaAct on two R2R unseen episodes. In the first example, SpaAct makes the correct turn at the first key intersection and stays aligned with the intended route, while the baseline misses this turn, follows the wrong path, and eventually fails. In the second example, SpaAct executes the early turns more precisely, enters the target region and successfully stops in front of the refrigerator. By contrast, the baseline deviates after the initial maneuver, misses the correct route into the target area. These examples show that SpaAct captures action-conditioned transition dynamics better and makes more reliable decisions at critical turning points in visually complex scenes.

\section{Conclusion}

In this paper, we present SpaAct, a transition-aware training framework for 
activating the dynamic spatial awareness in VLM-based vision-and-language navigation.
By introducing two spatial activation tasks, Action Retrospection for backward action reasoning and Future Frame Selection for forward transition prediction, SpaAct activates the navigation-relevant reasoning capabilities of VLMs. To further stabilize optimization, we propose TriPA, a tri-factor progressive curriculum learning strategy that organizes training samples from easy to hard according to trajectory, instruction, and motion complexity. Extensive experiments on standard VLN-CE benchmarks show that SpaAct consistently improves strong VLM-based baselines and achieves state-of-the-art performance. One limitation of the current study is the sim-to-real gap, especially when the camera setup and robot embodiment in simulation do not fully match those in real-world deployment. 
In future work, we will explore more realistic sensor and embodiment modeling to reduce this gap and further improve real-robot robustness.

\clearpage
\begin{acks}
  This work was supported by Fundamental and Interdisciplinary Disciplines Breakthrough Plan of the Ministry of Education of China under Grant JYB2025XDXM504.
\end{acks}

\bibliographystyle{ACM-Reference-Format}
\balance
\bibliography{sample-base}  

\end{document}